\documentclass[lettersize,journal]{IEEEtran}
\usepackage{amsmath,amsfonts}
\usepackage{algorithmic}
\usepackage{algorithm}
\usepackage{array}
\usepackage[caption=false,font=normalsize,labelfont=sf,textfont=sf]{subfig}
\usepackage{textcomp}
\usepackage{stfloats}
\usepackage{url}
\usepackage{verbatim}
\usepackage{graphicx}
\usepackage{cite}
\usepackage[table,xcdraw]{xcolor}
\usepackage{booktabs}
\usepackage{pifont} 
\usepackage{amssymb} 
\usepackage{multirow} 
\usepackage{float}
\usepackage{tcolorbox}
\tcbuselibrary{skins, breakable}
\usepackage[T1]{fontenc}

\begin{document}

\title{Transforming External Knowledge into Triplets for Enhanced Retrieval in RAG of LLMs}

\author{Xudong Wang, Chaoning Zhang*,~\IEEEmembership{Senior Member, IEEE}, Qigan Sun, Zhenzhen Huang, Chang Lu, Sheng Zheng, Zeyu Ma, Caiyan Qin, Yang Yang,~\IEEEmembership{Senior Member, IEEE}, Hengtao Shen,~\IEEEmembership{Fellow, IEEE}

\thanks{Xudong Wang and Qigan Sun are with the School of Computing, 
Kyung Hee University, Yongin-si, South Korea (e-mail: wl200203@khu.ac.kr; sunqigan0206@gmail.com).
Chaoning Zhang, Chang Lu, Sheng Zheng, Zeyu Ma, and Yang Yang are with School of Computer Science and Engineering, University of Electronic Science and Technology of China, Chengdu, China (email: chaoningzhang1990@gmail.com; george.changlu@gmail.com; zszhx2021@gmail.com; mazeyu@uestc.edu.cn; yang.yang@uestc.edu.cn).
Zhenzhen Huang is with School of Information and Software Engineering, University of Electronic Science and Technology of China, Chengdu, China (email: alley10086@gmail.com).
Caiyan Qin is with School of Robotics and Advanced Manufacture, Harbin Institute of Technology, Shenzhen, China (email: qincaiyan@hit.edu.cn). 
Hengtao Shen is with School of Computer Science and Technology, Tongji University, Shanghai, China (e-mail: shenhengtao@hotmail.com).}
\thanks{* Corresponding Author}
}

\maketitle

\begin{abstract}

Retrieval-Augmented Generation (RAG) mitigates hallucination in large language models (LLMs) by incorporating external knowledge during generation. However, the effectiveness of RAG depends not only on the design of the retriever and the capacity of the underlying model, but also on how retrieved evidence is structured and aligned with the query. Existing RAG approaches typically retrieve and concatenate unstructured text fragments as context, which often introduces redundant or weakly relevant information. This practice leads to excessive context accumulation, reduced semantic alignment, and fragmented reasoning chains, thereby degrading generation quality while increasing token consumption. To address these challenges, we propose Tri-RAG, a structured triplet-based retrieval framework that improves retrieval efficiency through reasoning-aligned context construction. Tri-RAG automatically transforms external knowledge from natural language into standardized structured triplets consisting of Condition, Proof, and Conclusion, explicitly capturing logical relations among knowledge fragments using lightweight prompt-based adaptation with frozen model parameters. Building on this representation, the triplet head Condition is treated as an explicit semantic anchor for retrieval and matching, enabling precise identification of query-relevant knowledge units without directly concatenating lengthy raw texts. As a result, Tri-RAG achieves a favorable balance between retrieval accuracy and context token efficiency. Experimental results across multiple benchmark datasets demonstrate that Tri-RAG significantly improves retrieval quality and reasoning efficiency, while producing more stable generation behavior and more efficient resource utilization in complex reasoning scenarios.

\end{abstract}


\section{Introduction}
Retrieval-Augmented Generation (RAG) equips large language models (LLMs) with access to external knowledge, improving factual grounding and supporting reasoning-intensive applications such as multi-hop question answering, knowledge-intensive retrieval, and structured decision support \cite{dixit2024sbi,levonian2023retrieval,zhao2024retrieval,li2024automated}. While modern LLMs are capable of generating fluent and context-aware responses, their performance in RAG settings is shaped not only by the retriever and the backbone model, but also by how retrieved evidence is organized and presented to the generator \cite{zhang2026lightweight}. In most existing pipelines, external knowledge is retrieved in the form of raw text segments, where semantic relations and logical dependencies remain implicit rather than explicitly structured. This mismatch between query form and evidence organization can weaken query--evidence alignment and reduce the reliability of downstream reasoning \cite{wu2022autoformalization,ferreira2020premise,chen2023theoremqa,yang2024formal,zhang2026learning,zhang2026tda}.

\begin{figure}
    \centering
    \includegraphics[width=1\linewidth]{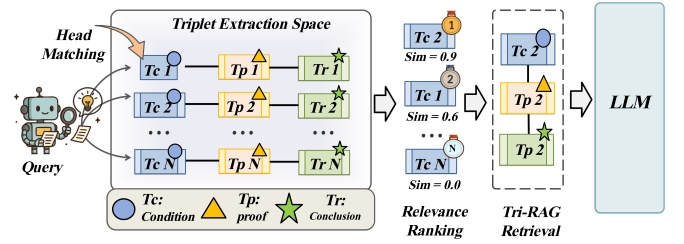}
    \caption{Triplet Matching for Efficient Reasoning. The input query is mapped to a condition representation that is matched against triplet heads in the knowledge base. The matched triplet (Condition, Proof, Conclusion) is returned to support structured reasoning with reduced unnecessary context expansion.}
    \label{fig:1}
\end{figure}

A common RAG workflow retrieves top-$k$ passages and concatenates them into a single context window. This fragment-level retrieval unit can introduce redundancy, dilute salient evidence with weakly related content, and yield contexts that are long yet semantically diffuse. The resulting context construction may hinder coherent reasoning and increase token usage, which can further amplify computation and latency costs \cite{zhang2026text,zheng2026llava,cao2026language,zheng2025joint}. These limitations motivate a representation-aware view of RAG: improving retrieval is not only a matter of ranking documents, but also a matter of organizing knowledge into units that better match the structure of queries and the needs of reasoning.

Prior work has explored structured processing of external knowledge to improve its usability within RAG. Representative directions include template-based extraction, schema-guided information structuring, and graph-based representations that organize entities and relations for downstream retrieval and reasoning. Such approaches can be effective in constrained settings, but they frequently rely on heuristic rules, domain-specific assumptions, or manual annotation, which complicates deployment under diverse natural language expressions and open-domain corpora\cite{amini2019mathqa,gao2023retrieval}. Another line of research leverages formal representations and verification systems \cite{de2015lean, bancerek2018role, barras1997coq, xu2026selfcorrectingrag} to provide precise semantics and proof checking.  These systems offer strong guarantees, yet their formal languages differ substantially from natural language, and building formalized resources typically requires substantial expert effort. This gap makes direct integration with large-scale RAG pipelines challenging, especially for knowledge that is expressed in varied, informal, or application-specific language.

To address these challenges, we propose  Tri-RAG, a structured knowledge transformation retrieval framework tailored for LLM-based RAG. Tri-RAG converts external knowledge written in natural language into standardized triplets of \textit{Condition}, \textit{Proof}, and \textit{Conclusion}, enabling retrieval over semantically focused units that expose key relational structure. The transformation is learned via soft prompt tuning \cite{lester2021power}: with the LLM backbone frozen, a small set of trainable prompt vectors guides the model to extract explicit triplet representations from unstructured knowledge sources. During retrieval, Tri-RAG treats the triplet head (Condition) as a dynamic semantic anchor for matching and returns the full triplet to support inference. This design aims to reduce ineffective concatenation of lengthy raw texts while preserving the information required for downstream reasoning.

We conduct empirical evaluations of Tri-RAG across multiple reasoning tasks. The results indicate that Tri-RAG improves retrieval quality and reasoning efficiency, and supports more stable generation under challenging retrieval conditions.

The main contributions of this work are as follows:
\begin{itemize}
    \item Tri-RAG is introduced as a structured knowledge transformation retrieval framework, representing external knowledge as triplets (Condition, Proof, Conclusion) to enhance retrieval and reasoning in RAG.
    \item A soft prompt tuning approach for triplet extraction is presented, which updates only a small set of prompt parameters while keeping the LLM backbone frozen, ensuring lightweight and scalable knowledge structuring.
    \item Tri-RAG is evaluated against representative RAG baselines across multiple reasoning benchmarks, demonstrating that triplet-based retrieval improves both retrieval quality and end-task performance.
\end{itemize}

\section{Related Work}

\subsection{RAG in Large Language Models}
Large Language Models (LLMs) exhibit strong reasoning and generation capabilities but often suffer from outdated internal knowledge and limited factual recall, especially in tasks requiring precise and up-to-date external information \cite{roberts2020much}. Retrieval-Augmented Generation (RAG) addresses this limitation by equipping LLMs with access to external knowledge sources \cite{lewis2020retrieval, guu2020retrieval, borgeaud2022improving}. A typical RAG pipeline retrieves top-$k$ relevant documents using dense encoders \cite{karpukhin2020dense, johnson2019billion} or sparse models such as ColBERT \cite{khattab2020colbert}, and concatenates the retrieved text with the query for generation \cite{izacard2020leveraging}. However, using unstructured text fragments as retrieval units often introduces redundancy, weak semantic alignment, and irrelevant content, which can degrade downstream reasoning quality \cite{pan2024unifying, soudani2024fine}.

Recent work improves retrieval through reranking \cite{jiang2023active, mao2021rider}, multi-document fusion \cite{zhang2025knowpo}, and dynamic retrieval strategies \cite{shi2023replug, sutton1998reinforcement}. Nevertheless, these efforts mainly focus on retrieval algorithms, while the structural representation and granularity of retrieved content remain largely unaddressed. More fundamentally, the lack of explicit structural modeling hinders semantic alignment between queries and evidence \cite{mialon2023augmented}. In reasoning-intensive tasks such as theorem proving or multi-hop question answering, plain-text retrieval often does not explicitly expose logical elements such as assumptions, conclusions, and intermediate steps, making downstream reasoning less stable or harder to control \cite{chai2025graphllm,zheng2026towards}.

Beyond text-centric retrieval, related research explores structured reasoning over knowledge graphs and relational data. Prior work investigates few-shot and low-resource reasoning over knowledge graphs \cite{yan2024multi}, structured graph or hypergraph modeling for complex learning systems \cite{li2024edugraph}, and reliable structure learning for causal reasoning \cite{yang2023learning}. While these approaches highlight the benefits of explicit structure, they typically assume pre-existing structured inputs, whereas Tri-RAG focuses on transforming unstructured natural language knowledge into retrieval-ready structured evidence.

\subsection{Soft Prompt Tuning for Large Language Models}
Prompt-based learning has emerged as an efficient alternative to full-parameter fine-tuning for adapting LLMs. Soft Prompt Tuning introduces a small set of trainable continuous prompt vectors prepended to the input sequence while keeping all model parameters frozen. By optimizing only prompt parameters, this approach significantly reduces training cost and memory overhead, and has demonstrated strong performance and generalization across models such as BERT \cite{li2021prefix}, GPT \cite{liu2024gpt}, and T5 \cite{wang2025parameter}. Subsequent work further extends this paradigm to improve expressiveness and adaptability. P-Tuning \cite{liu2021p} combines continuous prompts with discrete templates, while Prefix-Tuning \cite{li2021prefix} injects trainable prefix vectors into Transformer layers to better capture task-specific semantics. Recent studies also show that prompt-based adaptation can be effectively extended to specialized multimodal scenarios \cite{sun2026grasp}. Together, these advances establish soft prompt tuning as a practical and scalable strategy for controlling LLM behavior under parameter-efficient constraints.

However, in many reasoning-intensive domains, knowledge expressed in natural language often contains implicit structural semantics, such as premises, intermediate evidence, and conclusions, that are difficult for existing models to reliably identify and utilize. Traditional structure extraction methods typically rely on manually designed rules or templates, which limit scalability and generalization across diverse natural language expressions. To address this challenge, we leverage soft prompt tuning to guide frozen LLMs toward generating standardized triplet representations (Condition, Proof, Conclusion) from unstructured text. By introducing a small set of trainable prompt vectors, the model can be steered to produce structured outputs without modifying backbone parameters, enabling efficient and scalable construction of structured knowledge resources.

\section{Method}
\begin{figure*}[h]
    \centering
    \includegraphics[width=0.9\linewidth]{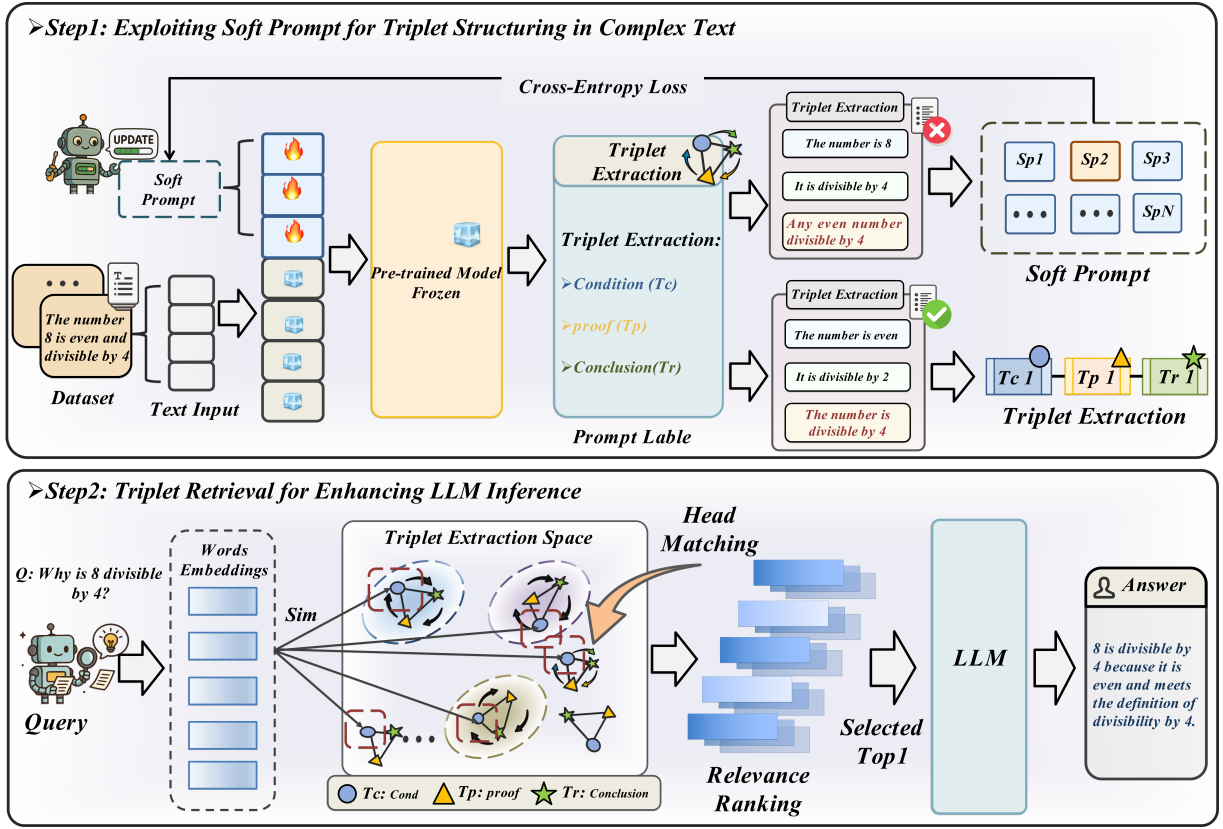}    
    \caption{Soft Prompt–Driven Triplet Structuring and Retrieval-Augmented Inference.
The upper part shows how soft prompts guide frozen LLMs to extract structured triplets (Condition, Proof, Conclusion). The lower part uses triplet retrieval to enhance LLM inference via relevance ranking and head matching.}
    \label{fig:2}
\end{figure*}

Tri-RAG is a structured knowledge-transformation retrieval framework designed for retrieval-augmented generation. The method is organized as follows: The Triplet Schema section (\ref{sec:Triplet}) explains how unstructured external knowledge is transformed into standardized triplets (Condition, Proof, Conclusion) with semantic constraints. The Triplet Extraction via Soft Prompt Tuning section (\ref{sec:Extraction}) details triplet extraction using soft prompt tuning. In the Triplet Knowledge Base Construction section (\ref{sec:kb}), a knowledge base indexed by triplet heads is built. Finally, the Inference: Condition-Anchored Retrieval and Full-Triplet Conditioning section (\ref{sec:inference}) describes condition-anchored retrieval and full triplet conditioning for generation. An overview of the pipeline is shown in Figure~\ref{fig:1} and Figure~\ref{fig:2}.

\subsection{Triplet Schema}
\label{sec:Triplet}
To improve the alignability and compositionality of retrieval units, Tri-RAG re-parameterizes external knowledge from free-form text into a structured triplet. Given a raw knowledge unit $x$ (e.g., a paragraph or passage), we map it to:
\begin{equation}
T=(T_c,\,T_p,\,T_r),
\end{equation}
where each field is defined with explicit semantic constraints:
\begin{itemize}
    \item \textbf{$T_c$ (Condition).} $T_c$ specifies the \emph{applicability scope} under which the claim holds, including entity definitions, contextual constraints (time/place/scenario), preconditions, assumptions, and input restrictions. $T_c$ is designed to provide the minimal context required for reliable query--knowledge alignment and serves as the \emph{triplet head} used for indexing and retrieval.
    \item \textbf{$T_p$ (Proof).} $T_p$ provides \emph{supporting rationale} connecting $T_c$ to $T_r$, such as a concise causal chain, rule-based justification, key intermediate steps, or evidence description. When the source text does not contain explicit rationale or when faithful extraction is not possible, we allow $T_p=\varnothing$ while preserving structural tags for consistent serialization.
    \item \textbf{$T_r$ (Conclusion).} $T_r$ states the \emph{core claim} that holds under $T_c$---a self-contained and verifiable proposition (e.g., a factual statement, rule outcome, operational result, or decision recommendation). $T_r$ avoids extraneous background not required for the claim, reducing noise and improving reuse across queries.
\end{itemize}

\paragraph{Interface to retrieval and generation}
Tri-RAG decouples the \emph{matching signal} from the \emph{generation evidence}: we retrieve by the triplet head $T_c$ to maximize alignment, and condition generation on the full triplet $(T_c,T_p,T_r)$ under a fixed context budget.

\paragraph{Granularity and compositionality}
Triplets are extracted at the same granularity as the downstream retrieval unit (e.g., paragraph/passage). If a raw unit contains multiple relatively independent claims, it can be decomposed into multiple triplets to reduce topic mixing and improve evidence compositionality. 
\paragraph{Problem Formulation}
Given a corpus of raw knowledge units $\mathcal{C}=\{x_j\}_{j=1}^{|\mathcal{C}|}$, Tri-RAG transforms each unit $x_j$ into a triplet $T_j=(T_{c,j},T_{p,j},T_{r,j})$ and builds a triplet knowledge base:
\begin{equation}
\mathcal{K}=\{T_j=(T_{c,j},T_{p,j},T_{r,j})\}_{j=1}^{|\mathcal{K}|}.
\end{equation}
At inference time, given a query $Q$, Tri-RAG retrieves candidate triplets by matching $Q$ against triplet heads $T_{c,j}$, and conditions generation on the retrieved full triplet(s) to produce the final answer $\hat{y}$ (details in Section~\ref{sec:inference}).

\subsection{Triplet Extraction via Soft Prompt Tuning}
\label{sec:Extraction}
To enable scalable tripletization without full-parameter fine-tuning, we adopt soft prompt tuning \cite{lester2021power} to train a structured extractor. Let $f_{\text{ext}}$ denote an open pre-trained Transformer used for triplet extraction, with $L$ layers and hidden size $d$. During training, we freeze all backbone parameters of $f_{\text{ext}}$ and optimize only a small set of continuous prompt embeddings.

\paragraph{Layer-wise soft prompts.}
Instead of using a single input-level prompt, we employ layer-wise (deep) soft prompts to increase extraction capacity while keeping the backbone frozen. 
Let $\mathcal{S}\subseteq\{1,\ldots,L\}$ denote the set of layers where prompts are injected, and let $L_p=|\mathcal{S}|$.
We introduce trainable prompt embeddings
\begin{equation}
\mathcal{E}_{\text{soft}}=\left\{\mathbf{E}^{(l)}_{\text{soft}} \in \mathbb{R}^{\ell \times d}\right\}_{l\in\mathcal{S}},
\end{equation}
where $\ell$ is the number of prompt tokens inserted per selected layer. 
The number of trainable parameters is therefore $\ell \cdot d \cdot L_p$.

For each raw knowledge unit $x^{(i)}$, we obtain the initial token embeddings
\begin{equation}
\mathbf{H}^{(0)}=\Phi\!\left(x^{(i)}\right),
\end{equation}
where $\Phi(\cdot)$ denotes tokenization and embedding lookup. At each Transformer layer $l$, we prepend the corresponding soft prompt to the hidden states:
\begin{equation}
\tilde{\mathbf{H}}^{(l-1)}=\left[\mathbf{E}^{(l)}_{\text{soft}};\ \mathbf{H}^{(l-1)}\right],
\end{equation}
and apply the frozen layer transformation
\begin{equation}
\tilde{\mathbf{H}}^{(l)}=\mathrm{Trans}^{(l)}\!\left(\tilde{\mathbf{H}}^{(l-1)}\right).
\end{equation}
To keep the sequence length stable across layers, we discard the prompt positions after each layer and retain only the hidden states corresponding to the original tokens:
\begin{equation}
\mathbf{H}^{(l)}=\tilde{\mathbf{H}}^{(l)}[\ell+1:\ ].
\end{equation}

\paragraph{Structured generation target.}
The extractor is trained to generate a fixed structured template:
\begin{equation}
y^{(i)}=\operatorname{Fmt}\!\left(T_c^{(i)},T_p^{(i)},T_r^{(i)}\right),
\end{equation}
where $\operatorname{Fmt}(\cdot)$ serializes a triplet into a tagged string
$\texttt{[C]}~T_c~\texttt{[P]}~T_p~\texttt{[R]}~T_r$.
When the source text does not contain explicit rationale, the model may output an empty or minimal $T_p^{(i)}$, which is later normalized during knowledge base construction (Section~\ref{sec:kb}).

\paragraph{Supervision construction.}
We optimize the layer-wise soft prompts $\mathcal{E}_{\text{soft}}$ on
\begin{equation}
\mathcal{D}_{\text{ext}}=\left\{\left(x^{(i)},T_c^{(i)},T_p^{(i)},T_r^{(i)}\right)\right\}_{i=1}^{N}.
\end{equation}
If human-annotated triplets are available, we use them directly. Otherwise, we construct weak supervision aligned with the downstream corpus: we derive $T_c$ targets from evidence annotations as retrieval-oriented semantic anchors, extract short rationales as $T_p$ from explanatory sentences or evidence summaries when available, and define $T_r$ as the core claim stated in the unit. We use a held-out validation split for model selection (and early stopping when applicable), and filter malformed or degenerate outputs during knowledge base construction (Section~\ref{sec:kb}).

\paragraph{Training objective}
Let $y^{(i)}$ denote the target formatted sequence. With layer-wise soft prompts, the extractor defines
\begin{equation}
P_{\theta}\!\left(y^{(i)} \mid x^{(i)};\mathcal{E}_{\text{soft}}\right)
=
\prod_{t=1}^{|y^{(i)}|}
P_{\theta}\!\left(y^{(i)}_{t}\mid y^{(i)}_{<t}, x^{(i)};\mathcal{E}_{\text{soft}}\right),
\end{equation}
and we minimize token-level cross-entropy:
\begin{equation}
\mathcal{L}
=
-\frac{1}{N}
\sum_{i=1}^{N}
\sum_{t=1}^{|y^{(i)}|}
\log P_{\theta}\!\left(y^{(i)}_{t}\mid y^{(i)}_{<t}, x^{(i)};\mathcal{E}_{\text{soft}}\right).
\end{equation}
Importantly, gradients are computed \emph{only} with respect to $\mathcal{E}_{\text{soft}}$:
\begin{equation}
\nabla_{\mathcal{E}_{\text{soft}}}\mathcal{L},
\end{equation}
while all backbone parameters of $f_{\text{ext}}$ remain frozen.

\subsection{Triplet Knowledge Base Construction}
\label{sec:kb}
We construct a triplet knowledge base $\mathcal{K}$ from the extracted triplets for retrieval and downstream conditioning.

\paragraph{Unit definition}
We segment the raw corpus into knowledge units $\{x_j\}$ that match the downstream retrieval granularity. For each unit $x_j$, the extractor produces:
\begin{equation}
T_j=\left(T_{c,j},T_{p,j},T_{r,j}\right).
\end{equation}

\paragraph{Normalization}
We standardize tags and surface forms across fields and apply light text cleaning (e.g., removing redundant whitespace/punctuation and duplicated tags). We discard degenerate outputs (e.g., empty $T_{c,j}$ or $T_{r,j}$) and remove samples with abnormal field lengths based on predefined bounds. We also de-duplicate near-identical triplet heads (Conditions) to reduce index redundancy and improve retrieval stability.

\paragraph{Quality control}
We apply a rule-based validator requiring (i) a complete tagged structure and (ii) field lengths within preset ranges. Optionally, we apply a lightweight confidence filter to remove low-quality outputs. 


\paragraph{Indexing}
We construct the retrieval index exclusively over the triplet heads $T_{c,j}$. 
Specifically, each $T_{c,j}$ is encoded into a dense vector representation and inserted into a dense vector index for similarity-based retrieval at query time. 
Each index entry is uniquely associated with its corresponding full triplet 
$T_j = (T_{c,j}, T_{p,j}, T_{r,j})$. 
Upon retrieval, the complete triplet is returned as a structured evidence unit and used for subsequent generation conditioning.

\begin{algorithm}[t]
\caption{Layer-wise Soft Prompt Tuning for Triplet Extraction}
\label{alg:softprompt}
\begin{algorithmic}[1]
\STATE \textbf{Input:} Supervision set $\mathcal{D}_{\text{ext}}=\{(x^{(i)},T_c^{(i)},T_p^{(i)},T_r^{(i)})\}_{i=1}^{N}$;
extractor $f_{\text{ext}}$ (frozen);
layer-wise soft prompts $\mathcal{E}_{\text{soft}}=\{\mathbf{E}^{(l)}_{\text{soft}}\}_{l\in\mathcal{S}}$;
learning rate $\eta$; epochs $K$
\STATE \textbf{Output:} Optimized soft prompts $\mathcal{E}_{\text{soft}}^{\star}$
\STATE {\color{gray}\# Only $\mathcal{E}_{\text{soft}}$ is trainable; all parameters of $f_{\text{ext}}$ are frozen.}

\FOR{epoch $=1$ \TO $K$}
    \FOR{each sample $(x^{(i)},T_c^{(i)},T_p^{(i)},T_r^{(i)}) \in \mathcal{D}_{\text{ext}}$}
        \STATE $y^{(i)} \leftarrow \operatorname{Fmt}(T_c^{(i)},T_p^{(i)},T_r^{(i)})$
        \STATE {\color{gray}\# Forward pass with layer-wise prompt injection.}
        \STATE $\hat{y}^{(i)} \leftarrow f_{\text{ext}}\!\left(x^{(i)};\mathcal{E}_{\text{soft}}\right)$
        \STATE $\mathcal{L} \leftarrow \operatorname{CE}(\hat{y}^{(i)}, y^{(i)})$
        \STATE $\mathbf{g} \leftarrow \nabla_{\mathcal{E}_{\text{soft}}}\mathcal{L}$
        \STATE $\mathcal{E}_{\text{soft}} \leftarrow \mathcal{E}_{\text{soft}} - \eta\,\mathbf{g}$
    \ENDFOR
\ENDFOR
\STATE $\mathcal{E}_{\text{soft}}^{\star} \leftarrow \mathcal{E}_{\text{soft}}$
\STATE \textbf{Return:} $\mathcal{E}_{\text{soft}}^{\star}$
\end{algorithmic}
\end{algorithm}
\subsection{Inference: Condition-Anchored Retrieval and Full-Triplet Conditioning}
\label{sec:inference}

Given a query $Q$, Tri-RAG follows a unified two-stage inference protocol: retrieval is driven solely by the triplet head $T_c$, while generation is conditioned on the full triplet $(T_c, T_p, T_r)$. 
Let $k$ denote the retrieval depth, $m$ the number of triplets selected for generation conditioning ($m \leq k$), and $B$ the token budget allocated to the evidence block. 
The overall inference procedure is summarized in Algorithm~\ref{alg:trirag}.

\paragraph{Condition-Anchored Retrieval}
During retrieval, Tri-RAG uses only the triplet head $T_c$ as the matching unit. 
Specifically, the query $Q$ and each triplet head $T_{c,j}$ are encoded into a shared vector space using the same retrieval encoder $\mathrm{Enc}_R(\cdot)$:
\begin{equation}
\mathbf{q} = \mathrm{Enc}_R(Q), \qquad \mathbf{k}_j = \mathrm{Enc}_R(T_{c,j}).
\end{equation}
Candidate triplets are retrieved via similarity search over the triplet knowledge base $\mathcal{K}$:
\begin{equation}
\mathcal{K}_{\text{top-}k} =
\mathrm{TopK}\!\left(
\mathbf{q}, \{\mathbf{k}_j\}_{j=1}^{|\mathcal{K}|}
\right).
\end{equation}
All candidates are ranked according to their similarity scores between the query and triplet heads, and the top $m$ triplets are selected for downstream generation conditioning. 
No additional reranking module is introduced, ensuring a simple and controlled retrieval pipeline.

\paragraph{Evidence Construction and Full-Triplet Conditioning}
For generation, the selected $m$ triplets are serialized using a fixed template and concatenated into an evidence block $C$.
Each triplet is represented in a structured textual form with explicit field tags, namely
\texttt{[C]}~$T_c$, \texttt{[P]}~$T_p$, and \texttt{[R]}~$T_r$, corresponding to the condition, rationale, and conclusion fields, respectively. During evidence construction, we enforce a unified token budget $B$ measured using the tokenizer of the generation model.
If the concatenated evidence exceeds the budget, truncation is applied strictly within field boundaries to preserve structural integrity.
By default, truncation is applied to the rationale field $T_p$ first, in order to maximize effective information density under a fixed context length constraint while preserving alignment-critical conditions and core conclusions. The generator input and output are defined as:

\begin{equation}
\mathbf{x} = [Q; C], \qquad \hat{y} = f_{\mathrm{gen}}(\mathbf{x}),
\end{equation}
where $f_{\mathrm{gen}}$ denotes the language model used for answer generation.

\begin{algorithm}[t]
\caption{Tri-RAG Inference}
\label{alg:trirag}
\begin{algorithmic}[1]
\STATE \textbf{Input:} Query $Q$; triplet KB $\mathcal{K}=\{(T_{c,j},T_{p,j},T_{r,j})\}_{j=1}^{|\mathcal{K}|}$; retrieval encoder $\mathrm{Enc}_R$; retrieval depth $k$; selected triplets $m$ ($m \leq k$); token budget $B$; generator $f_{\mathrm{gen}}$
\STATE \textbf{Output:} Answer $\hat{y}$

\STATE $\mathbf{q} \leftarrow \mathrm{Enc}_R(Q)$
\STATE {\color{gray}\# Retrieval is performed on triplet heads ($T_c$) only.}
\STATE $\mathcal{K}_{\text{top-}k} \leftarrow
\mathrm{TopK}\!\left(\mathbf{q}, \{\mathrm{Enc}_R(T_{c,j})\}_{j=1}^{|\mathcal{K}|}\right)$

\STATE Select the top $m$ triplets $\mathcal{T}$ by similarity score

\STATE {\color{gray}\# Full triplets are used as generation evidence.}
\STATE $C \leftarrow \mathrm{Format}(\mathcal{T}; B)$

\STATE {\color{gray}\# Enforce token budget with field-level truncation (truncate $T_p$ first).}
\STATE $\mathbf{x} \leftarrow [Q; C]$

\STATE $\hat{y} \leftarrow f_{\mathrm{gen}}(\mathbf{x})$
\STATE \textbf{Return:} $\hat{y}$
\end{algorithmic}
\end{algorithm}

\paragraph{Model Usage}
Soft prompt tuning requires trainable prompt embeddings; therefore, triplet extraction is performed using an open, trainable language model.
The generation model $f_{\mathrm{gen}}$ may be either an open-source model or a closed API-based LLM, but it is not involved in training or updating the soft prompts.

\section{Experiment}

\subsection{Experimental Setup}
\label{sec:exp_setup}

\subsubsection{Datasets}
\label{sec:datasets}

To comprehensively evaluate the effectiveness of Tri-RAG in long-context question answering and multi-hop reasoning, we conduct experiments along two complementary axes:Long-context and multi-hop QA benchmarks, and HotpotQA Robustness Variants.

\paragraph{Long-context and multi-hop QA benchmarks}
\label{sec:Long}
We evaluate end-to-end answering performance on LongBench\cite{bai2023longbench} and a set of widely used QA benchmarks, including HotpotQA\cite{yang2018hotpotqa}, 2WikiMultihopQA\cite{ho2020constructing}, MuSiQue\cite{trivedi2022musique}, Natural Questions (NQ)\cite{kwiatkowski2019natural}, and SQuAD\cite{rajpurkar2016squad}. Collectively, these datasets span single-hop factual QA, compositional multi-hop reasoning, and reading comprehension with varying context lengths, enabling assessment of generalization across different reasoning depths and evidence aggregation patterns.

\paragraph{HotpotQA Robustness Variants.}
We evaluate robustness on four variants derived from HotpotQA~\cite{yang2018hotpotqa} that cover different retrieval scales and prior-knowledge conditions:
\begin{itemize}
    \item Hotpot-Dist (HotpotQA-Distractor): each query is paired with a small candidate set (typically 10 documents) containing supporting and distractor documents.
    \item Hotpot-Full (HotpotQA-Fullwiki): evidence is retrieved from the full Wikipedia corpus.
    \item Shuffle-Hotpot-Dist: an entity-replacement robustness variant introduced by Zhu et al.~\cite{zhu2025knowledge}, where entities in questions and contexts are replaced with type-consistent alternatives to suppress memorized factual priors, forcing reliance on retrieved evidence.
    \item Shuffle-Hotpot-Full: the entity-replacement counterpart of Hotpot-Full with the same substitution strategy at full-corpus scale.
\end{itemize}

\subsubsection{Baselines Adopted}
\label{sec:baselines}

We design baselines from three perspectives: search and reasoning strategies, RAG framework baselines, and parameter-efficient adaptation settings. This design aims to make the source of performance gains attributable and interpretable.

\paragraph{Search and reasoning strategies}
We evaluate Tri-RAG under four representative retrieval--reasoning strategies used as the ``Method'' dimension: Direct, ReAct~\cite{yao2023react}, Self-Act, and Aflow. Direct performs single-pass generation without explicit planning, while ReAct interleaves reasoning and actions across multiple steps. Self-Act focuses on self-initiated action selection with iterative evidence completion, and Aflow follows a staged action-flow procedure for evidence aggregation. All strategies are evaluated under matched inference settings; Tri-RAG differs only in evidence representation and the use of the Condition field as the retrieval key.

\paragraph{RAG framework baselines}
To compare Tri-RAG against established retrieval-augmented systems in terms of retrieval quality and answer quality, we adopt the following framework-level baselines:
\begin{itemize}
    \item LLM-only: no retrieval; the model answers solely based on parametric knowledge.
    \item Semantic RAG\cite{jiang2023active}: dense semantic retrieval over unstructured passages, used as context for generation.
    \item Hybrid RAG\cite{gao2021complement}: hybrid sparse and dense retrieval to improve coverage and relevance.
    \item LightRAG\cite{guo2024lightrag}: a simplified RAG design optimized for low latency and lightweight deployment.
    \item GraphRAG\cite{edge2024local}: organizes knowledge in graph structures to support entity/relation chaining in multi-hop reasoning.
    \item KG$^2$RAG\cite{zhu2025knowledge}: extends graph-based retrieval by coupling knowledge-graph reasoning with the RAG pipeline to improve grounding.
    \item Tri-RAG: transforms external knowledge into a triplet representation (Condition, Proof, Conclusion), retrieves by matching Condition as a semantic anchor, and conditions generation on the full triplet evidence.
\end{itemize}

\paragraph{Parameter-efficient adaptation baselines.}
Tri-RAG depends on a parameter-efficient extractor to produce triplet-structured evidence at low adaptation cost. To ensure a fair comparison under a constrained trainable-parameter budget, we evaluate several parameter-efficient adaptation schemes for the extractor while keeping the backbone model frozen whenever applicable:
\begin{itemize}
    \item Pretrained: using the frozen backbone directly without any task-specific adaptation.
    \item Prefix tuning \cite{li2021prefix}: prepending a small number of trainable continuous prefix vectors to the input sequence, while leaving all backbone parameters unchanged.
    \item LoRA \cite{hu2022lora}: inserting low-rank adaptation modules into selected layers to enable efficient fine-tuning with a limited number of trainable parameters.
    \item Tri-RAG (Ours): our extractor configuration, evaluated end-to-end against the above baselines on the same datasets to assess both the effectiveness and parameter efficiency of structured triplet extraction.
\end{itemize}

\subsubsection{Experimental Details}
\label{sec:exp_details}

To ensure fairness and reproducibility, we decompose the system into two components: the generator module and the retrieval--structuring module (Retriever + Triplet KB). Key variables are controlled across methods so that performance differences primarily reflect retrieval behavior and evidence organization rather than generation-side configurations.

\paragraph{Generator backbones.}
We evaluate end-to-end performance using multiple generator backbones, including GPT-4o-mini, GPT-4.1-nano, Qwen-Long, Qwen-Max, and Gemma-2-27B. For each dataset, all methods share the same prompt template, maximum output length, and decoding configuration, ensuring that comparisons isolate the effect of retrieval and evidence construction.

\paragraph{Triplet extraction backbones.}
Since triplet extraction involves trainable parameters, the structured-output extractor is trained on open-weight backbones. We use DeepSeek-7B, Llama-3-8B, and Qwen3-8B, following their official tokenization and preprocessing pipelines. During training, backbone parameters are frozen and only parameter-efficient components (e.g., soft prompts) are optimized, yielding a lightweight extractor for Triplet KB construction.

\paragraph{Training setup.}
All experiments are conducted on a single NVIDIA RTX 4090 GPU. We use AdamW with an initial learning rate of $1\times10^{-4}$ and weight decay of 0.01. The learning rate is linearly warmed up over the first 10\% of training steps and then linearly decayed. The batch size is 6, with up to 20 epochs and early stopping based on validation performance (patience = 5). 

\paragraph{Retrieval and evaluation.}
All methods share the same retrieval encoder, index implementation, retrieval depth $k$, and evidence token budget $B$. 
When the budget is exceeded, Tri-RAG applies field-aware truncation that prioritizes the Condition and Conclusion fields, whereas passage-based baselines use passage-level truncation. 
Generation is performed with identical settings across methods. 
End-to-end QA performance is evaluated using standard exact-match or F1 metrics, and structured extraction quality is assessed via a unified LLM-as-a-Judge protocol (Section~\ref{sec:Structured Extraction and Efficiency}).

\begin{table*}[t!]
\centering
\caption{Performance of search strategies (Direct, ReAct, Self-Act, Aflow, and Tri-RAG) across five QA benchmarks. All metrics are reported as F1 scores (\%).}
\resizebox{0.8\linewidth}{!}{
\setlength{\tabcolsep}{3pt}
\renewcommand{\arraystretch}{0.8}
\begin{tabular}{clcccccc}
\toprule
\textbf{Model} & \textbf{Method} & \textbf{LongBench} & \textbf{2WikiMultihopQA} & \textbf{MuSiQue} & \textbf{NQ} & \textbf{SQuAD} & \textbf{AVG} \\
\midrule
\multirow{5}{*}{GPT-4o-mini}
 & Direct        & 58.1 & 48.9 & 40.1 & 36.4 & 51.2 & 46.9 \\
 & ReAct         & 62.7 & 52.3 & 53.6 & 48.9 & 55.3 & 54.6 \\
 & Self-Act      & 68.9 & 69.2 & 67.3 & 62.4 & 69.5 & 67.5 \\
 & Aflow         & 61.0 & 65.7 & 71.6 & 67.9 & 73.7 & 68.0 \\
 \rowcolor{gray!15} & \textbf{Tri-RAG} & \textbf{70.9} & \textbf{73.3} & \textbf{74.6} & \textbf{69.8} & \textbf{75.3} & \textbf{72.8} \\
\cmidrule(lr){1-8}

\multirow{5}{*}{GPT-4.1-nano}
 & Direct        & 56.3 & 45.3 & 38.2 & 33.7 & 49.3 & 44.6 \\
 & ReAct         & 59.7 & 48.9 & 47.3 & 45.4 & 50.9 & 50.4 \\
 & Self-Act      & 64.5 & 62.2 & 61.1 & 59.8 & 62.7 & 62.1 \\
 & Aflow         & 66.1 & 70.1 & 69.4 & 66.5 & 72.1 & 68.8 \\
 \rowcolor{gray!15} & \textbf{Tri-RAG} & \textbf{68.2} & \textbf{73.5} & \textbf{71.2} & \textbf{69.8} & \textbf{77.1} & \textbf{72.0} \\
\cmidrule(lr){1-8}

\multirow{5}{*}{Qwen-Long}
 & Direct        & 59.3 & 51.9 & 55.4 & 43.4 & 54.3 & 52.9 \\
 & ReAct         & 64.2 & 57.2 & 58.9 & 56.3 & 58.7 & 59.1 \\
 & Self-Act      & 68.3 & 73.1 & 72.3 & 68.9 & 70.4 & 70.6 \\
 & Aflow         & 70.8 & 76.8 & 78.9 & 74.6 & 77.1 & 75.6 \\
 \rowcolor{gray!15} & \textbf{Tri-RAG} & \textbf{73.5} & \textbf{78.3} & \textbf{81.3} & \textbf{78.7} & \textbf{79.2} & \textbf{78.2} \\
\cmidrule(lr){1-8}

\multirow{5}{*}{Qwen-Max}
 & Direct        & 47.2 & 46.6 & 49.9 & 53.3 & 48.5 & 49.1 \\
 & ReAct         & 52.3 & 53.1 & 53.6 & 59.8 & 54.5 & 54.7 \\
 & Self-Act      & 58.7 & 64.8 & 61.8 & 67.8 & 68.3 & 64.3 \\
 & Aflow         & 61.2 & 68.1 & 65.3 & 73.9 & 74.0 & 68.5 \\
 \rowcolor{gray!15} & \textbf{Tri-RAG} & \textbf{62.8} & \textbf{70.9} & \textbf{68.3} & \textbf{78.8} & \textbf{76.5} & \textbf{71.5} \\
\cmidrule(lr){1-8}

\multirow{5}{*}{Gemma-2-27B}
 & Direct        & 49.3 & 43.4 & 48.4 & 51.3 & 48.5 & 48.2 \\
 & ReAct         & 52.5 & 50.8 & 51.7 & 57.8 & 54.5 & 53.5 \\
 & Self-Act      & 56.4 & 59.7 & 59.7 & 65.3 & 65.7 & 61.4 \\
 & Aflow         & 60.8 & 63.2 & 63.4 & 69.4 & 71.9 & 65.7 \\
 \rowcolor{gray!15} & \textbf{Tri-RAG} & \textbf{63.3} & \textbf{67.9} & \textbf{67.1} & \textbf{72.5} & \textbf{75.8} & \textbf{69.3} \\
\bottomrule
\end{tabular}
}
\label{tab:1}
\end{table*}

\subsection{Performance Evaluation}

In this section, we evaluate Tri-RAG on long-context question answering and multi-hop reasoning, focusing on effectiveness, reliability, and robustness to contextual noise. The evaluation spans three dimensions: end-to-end answer quality, retrieval quality, and robustness under perturbations. We examine whether the gains of Tri-RAG persist across diverse generator backbones and representative search or reasoning strategies, mitigating reliance on specific model families or inference heuristics. We further analyze a controlled multi-hop benchmark that allows retrieval and generation to be assessed separately, enabling a clearer connection between evidence selection and answer correctness.

Table~\ref{tab:1} summarizes end-to-end F1 scores on five QA benchmarks under multiple generator backbones, including GPT-4o-mini, GPT-4.1-nano, Qwen-Long, Qwen-Max, and Gemma-2-27B. To ensure methodological consistency, all methods use the same prompt template, output-length budget, and decoding configuration for each dataset; consequently, performance differences are attributable primarily to retrieval behavior and evidence organization rather than generation-side tuning. Across all backbones, Tri-RAG attains the strongest average performance and remains consistently competitive on each benchmark. For GPT-4o-mini, Tri-RAG improves the average F1 from 67.5\% to 72.8\% relative to the best competing strategy, while achieving 73.3\% on 2WikiMultihopQA and 74.6\% on MuSiQue. The trend persists for GPT-4.1-nano, where the average F1 increases from 68.8\% to 72.0\% and performance on 2WikiMultihopQA rises from 70.1\% to 73.5\%. Comparable gains are observed for open-weight long-context models. On Qwen-Long, Tri-RAG increases the average F1 from 75.6\% to 78.2\%; on Qwen-Max and Gemma-2-27B, the average F1 increases from 68.5\% to 71.5\% and from 65.7\% to 69.3\%, respectively. Collectively, these results indicate that the benefits of Tri-RAG are not contingent on a particular generator backbone or task distribution, and instead generalize across model capacities and benchmark characteristics.

End-to-end metrics, however, do not by themselves identify the source of improvement. We therefore perform a mechanism-oriented evaluation on HotpotQA, a standard multi-hop benchmark with annotated supporting evidence, which supports a direct assessment of both retrieval and answering. Table~\ref{tab:2} reports response quality and retrieval quality on four HotpotQA variants. Hotpot-Dist and Hotpot-Full correspond to the standard distractor and full-wiki settings, respectively. Shuffle-Hotpot-Dist and Shuffle-Hotpot-Full apply an entity-replacement perturbation, where entities in both questions and contexts are substituted with type-consistent alternatives. This perturbation weakens direct entity-specific cues and provides a controlled setting for assessing robustness under reduced entity-level priors.

\begin{table*}[t!]
\centering
\caption{Performance of different methods on HotpotQA datasets and their respective response and retrieval qualities.}
\resizebox{\linewidth}{!}{
\begin{tabular}{lcccccccccccc}
\toprule
\multirow{3}{*}{Methods} & \multicolumn{3}{c}{Hotpot-Dist} & \multicolumn{3}{c}{Hotpot-Full} & \multicolumn{3}{c}{Shuffle-Hotpot-Dist} & \multicolumn{3}{c}{Shuffle-Hotpot-Full} \\
\cmidrule(lr){2-4} \cmidrule(lr){5-7} \cmidrule(lr){8-10} \cmidrule(lr){11-13}
& F1 & Precision & Recall & F1 & Precision & Recall & F1 & Precision & Recall & F1 & Precision & Recall \\
\hline
\multicolumn{13}{c}{\textbf{Response Quality}} \\
LLM-only & 23.7 & 25.9 & 23.4 & 23.7 & 25.9 & 23.4 & 15.8 & 17.5 & 15.8 & 15.8 & 17.5 & 15.8 \\
Semantic RAG & 61.7 & 64.6 & 64.3 & 52.8 & 55.8 & 53.5 & 50.8 & 53.3 & 52.4 & 42.2 & 44.9 & 43.3 \\
+ Rerank & 65.2 & 68.5 & 66.5 & 58.7 & 61.3 & 60.3 & 53.2 & 56.0 & 54.6 & 44.7 & 47.6 & 45.6 \\
Hybrid RAG & 65.3 & 67.6 & 65.5 & 55.1 & 58.2 & 55.8 & 52.0 & 54.8 & 53.4 & 44.3 & 47.3 & 44.6 \\
LightRAG & 29.3 & 28.8 & 48.0 & 26.1 & 25.9 & 36.4 & 28.5 & 28.4 & 40.4 & 20.2 & 19.9 & 29.3 \\
GraphRAG & 40.0 & 40.8 & 49.1 & 16.9 & 15.7 & 42.9 & 35.1 & 36.5 & 40.1 & 16.3 & 15.5 & 36.2 \\
KG\textsuperscript{2}RAG & 66.3 & 69.0 & 68.3 & 63.1 & 66.5 & 64.3 & 54.5 & 57.2 & 56.6 & 50.7 & 53.9 & 51.2 \\
\rowcolor{gray!15}
\textbf{Tri-RAG(Ours)} & \textbf{67.8} & \textbf{69.5} & \textbf{69.3} & \textbf{64.2} & \textbf{67.8} & \textbf{64.8} & \textbf{55.9} & \textbf{58.5} & \textbf{57.8} & \textbf{51.7} & \textbf{55.6} & \textbf{52.4} \\
\hline
\multicolumn{13}{c}{\textbf{Retrieval Quality}} \\
Semantic RAG & 34.3 & 20.6 & 89.4 & 30.0 & 17.8 & 79.0 & 32.1 & 20.1 & 83.7 & 26.8 & 16.7 & 70.8 \\
+ Rerank & 35.7 & 22.4 & \underline{93.2} & 30.6 & 19.7 & 83.3 & 33.9 & 21.3 & \underline{88.6} & 28.6 & 17.9 & 75.4 \\
Hybrid RAG & 35.4 & 22.2 & 92.1 & 30.2 & 18.9 & 79.5 & 33.4 & 21.0 & 83.7 & 27.9 & 17.4 & 73.9 \\
LightRAG & 23.4 & 15.0 & 63.8 & 13.2 & 8.3 & 34.0 & 22.7 & 14.8 & 53.5 & 11.6 & 7.3 & 29.5 \\
GraphRAG & 25.5 & 16.7 & 59.4 & 18.0 & 11.3 & 47.0 & 21.0 & 13.8 & 48.2 & 19.9 & 12.6 & 51.0 \\
KG\textsuperscript{2}RAG & 43.6 & 30.1 & 90.8 & 31.0 & 20.3 & \underline{83.8} & 40.5 & 27.9 & 84.0 & 30.5 & 19.3 & 79.0 \\
\rowcolor{gray!15}
\textbf{Tri-RAG(Ours)} & \textbf{44.1} & \textbf{31.9} & 91.4 & \textbf{32.0} & \textbf{21.4} & 83.1 & \textbf{41.9} & \textbf{28.9} & 85.7 & \textbf{31.5} & \textbf{20.5} & \textbf{79.2} \\
\bottomrule
\end{tabular}
}
\label{tab:2}
\end{table*}

Tri-RAG yields consistent improvements in response quality in both standard and perturbed settings. On Hotpot-Dist, Tri-RAG achieves an answer F1 of 67.8\%, exceeding KG$^2$RAG at 66.3\% and Semantic RAG at 61.7\%. On Hotpot-Full, Tri-RAG reaches 64.2\%, again outperforming the strongest baseline at 63.1\%. Under perturbation, the advantage remains, with Tri-RAG attaining 55.9\% on Shuffle-Hotpot-Dist and 51.7\% on Shuffle-Hotpot-Full, compared with 54.5\% and 50.7\% for KG$^2$RAG. This indicates that Tri-RAG exhibits reduced sensitivity to contextual shuffling and maintains more stable performance in the presence of noisy or reordered evidence.

Retrieval-quality comparisons provide a direct account of this behavior. On Hotpot-Dist, Tri-RAG achieves retrieval precision of 31.9\% and retrieval F1 of 44.1\%, improving upon KG$^2$RAG at 30.1\% and 43.6\% while maintaining high recall at 91.4\%. On Shuffle-Hotpot-Dist, Tri-RAG remains superior in retrieval F1, with 41.9\% compared to 40.5\% for KG$^2$RAG. The combination of high recall with improved precision suggests that Tri-RAG does not rely on narrowing retrieval to gain apparent relevance; instead, it increases evidence purity while preserving coverage, thereby mitigating context-budget dilution and reducing distracting content during generation. These observations are consistent with the Tri-RAG design: Condition-anchored retrieval constrains matching and reduces drift, while full-triplet conditioning supplies structured, verifiable evidence for downstream reasoning. Taken together, Table~\ref{tab:1} establishes cross-backbone effectiveness under diverse inference strategies, and Table~\ref{tab:2} provides a diagnostic link from improved evidence selection to improved answer quality and robustness under perturbations.

\begin{table*}[t]
\centering
\caption{Comparison of system cost for structured knowledge construction and query-time inference.}
\label{tab:systemcost}
\renewcommand{\arraystretch}{1.3}
\setlength{\tabcolsep}{12pt}
\begin{tabular}{lcccccc}
\toprule
Method 
& \multicolumn{4}{c}{Knowledge construction} 
& \multicolumn{2}{c}{Inference} \\
\cmidrule(lr){2-5} \cmidrule(lr){6-7}
& Input tokens & Output tokens & LLM calls & Time 
& Retrieval & Generation \\
\midrule
LightRAG 
& 1,269 & 381 & 1 & 3\,s 
& 40\,ms & 5,600\,ms \\

GraphRAG 
& 2,791 & 629 & 5 & 6\,s 
& 42\,ms & 5,500\,ms \\

KG$^2$RAG 
& 561 & 22 & 1 & 1\,s 
& 25\,ms & 2,300\,ms \\

Tri-RAG (Ours) 
& 640 & 96 & 1 & 1\,s 
& 18\,ms & 1,900\,ms \\
\bottomrule
\end{tabular}
\end{table*}

\subsection{System Cost Analysis: From Construction to Inference}
\label{sec:system_cost}

To examine whether structured evidence introduces additional system overhead, we conduct a cost-diagnostic study that decomposes the pipeline into two stages: offline knowledge construction (transforming raw text into structured evidence units) and online inference (query-time retrieval and generation under a unified serving setting). Table~\ref{tab:systemcost} summarizes the results, enabling a direct assessment of whether Tri-RAG’s efficiency gains come at the expense of extra preprocessing cost or increased latency.

\paragraph{Offline construction.}
We report per-unit input/output tokens, the number of LLM calls, and the wall-clock time required to construct one knowledge unit. GraphRAG incurs the highest fixed cost, as it relies on multi-round LLM interactions to generate and complete graph structure (5 calls on average), resulting in 2,791/629 input/output tokens and about 6\,s per unit. LightRAG uses a single call but still requires long-context prompting, with 1,269/381 tokens per unit. In contrast, Tri-RAG constructs compact triplet units with a single call, requiring 640/96 tokens and about 1\,s per unit. Its construction cost is close to KG$^2$RAG (561/22 tokens; 1 call; about 1\,s), indicating that tripletization can be achieved with a controlled, lightweight preprocessing budget rather than a multi-round or long-context construction pipeline.

\paragraph{Online inference.}
We further measure average retrieval and generation time per query. Tri-RAG reduces retrieval latency to 18\,ms, compared with 25\,ms for KG$^2$RAG and 40--42\,ms for LightRAG/GraphRAG, consistent with using \textit{Condition} as a tight semantic key that shrinks the matching space and limits low-utility candidates. On the generation side, Tri-RAG achieves 1,900\,ms, improving over KG$^2$RAG (2,300\,ms; $\sim$17\% reduction) and substantially outperforming LightRAG/GraphRAG (5,600/5,500\,ms; $\sim$65\% reduction). This gap aligns with the evidence organization effect: triplet evidence provides clearer boundaries and higher information density, reducing the fraction of the context window occupied by redundant or weakly relevant fragments.

Overall, the study supports the interpretation that structure does not necessarily imply higher cost. When structured units expose an explicit retrieval anchor and maintain compact evidence boundaries, both offline construction (single-call, moderate token budgets) and online inference (lower retrieval and generation latency) can benefit concurrently, yielding a more deployable quality--efficiency trade-off for reasoning-intensive settings.

\subsection{Structured Extraction and Efficiency}
\label{sec:Structured Extraction and Efficiency}

\begin{table*}[t]
\centering
\caption{End-to-end F1 (\%) on multi-hop QA benchmarks for different extractor backbones under parameter-efficient tuning methods.}
\label{tab:3}
\renewcommand{\arraystretch}{1.3}
\setlength{\tabcolsep}{10pt}
\begin{tabular}{lccc|ccc|ccc|c}
\toprule
\multirow{2}{*}{Method}
& \multicolumn{3}{c|}{DeepSeek-7B} 
& \multicolumn{3}{c|}{Llama-3-8B} 
& \multicolumn{3}{c|}{Qwen3-8B} 
& \multirow{2}{*}{Avg.}  \\
\cmidrule(lr){2-4} \cmidrule(lr){5-7} \cmidrule(lr){8-10}
& Hotpot & 2Wiki & MuSiQue 
& Hotpot & 2Wiki & MuSiQue 
& Hotpot & 2Wiki & MuSiQue 
&  \\
\midrule
Pretrained 
& 49.8 & 46.5 & 44.1 
& 52.1 & 48.7 & 46.0 
& 53.5 & 50.2 & 48.4 
& 48.8 \\

Prefix Tuning 
& 56.7 & 52.8 & 50.3 
& 58.9 & 55.1 & 52.2 
& 60.1 & 58.4 & 55.6 
& 55.6 \\

LoRA
& 63.9 & 60.4 & 57.1 
& 66.2 & 62.9 & 59.5 
& 68.4 & 67.1 & 64.8 
& 63.4 \\

\rowcolor{gray!15}
\textbf{Tri-RAG (SP)} 
& \textbf{66.8} & \textbf{63.7} & \textbf{60.2} 
& \textbf{69.0} & \textbf{73.3} & \textbf{74.6} 
& \textbf{72.5} & \textbf{71.8} & \textbf{70.4} 
& \textbf{69.6} \\
\bottomrule
\end{tabular}
\end{table*}
Tri-RAG incorporates a structured pre-processing stage that compresses and reorganizes unstructured external knowledge units into retrievable and compositional triplet evidence. We therefore analyze its quality--cost trade-off along three complementary axes: the adaptation strategy used for structured extraction, the end-to-end downstream gains on multi-hop question answering, and the associated system overhead. Table \ref{tab:3} compares representative adaptation paradigms for triplet extraction on multiple open-weight backbones and evaluates how each choice affects downstream multi-hop QA. Across HotpotQA, 2WikiMultihopQA, and MuSiQue, Tri-RAG yields the best or consistently superior average performance. The trend is stable across backbones, indicating that effective structured extraction can be learned under a strictly parameter-efficient regime: the backbone remains frozen and only a soft prompt is optimized. Importantly, the improvements are not confined to extraction accuracy in isolation, but translate into measurable gains in retrieval-augmented reasoning end to end. This evidence reduces the plausibility of attributing the gains primarily to heavy parameter updates, and instead supports the interpretation that the dominant benefit comes from improved retrieval alignment and evidence organization enabled by the triplet representation.

\begin{table}[ht]
\centering
\caption{Triplet extraction quality (1--5, mean$\pm$std) and efficiency (trainable parameters) for Prefix Tuning, LoRA, and Tri-RAG.}
\label{tab:4}
\renewcommand{\arraystretch}{1.2}
\setlength{\tabcolsep}{4pt}
\resizebox{\columnwidth}{!}{
\begin{tabular}{lcccc}
\toprule
\textbf{Metric} & \textbf{Prefix Tuning } & \textbf{LoRA} & \textbf{Tri-RAG} \\ \midrule
\textit{\textbf{Quality Metrics}} & & & \\
Logical Faithfulness & 4.12 $\pm$ 0.35 & 4.41 $\pm$ 0.12 & \textbf{4.48 $\pm$ 0.18} \\
Structural Integrity & 3.75 $\pm$ 0.82 & 4.75 $\pm$ 0.10 & \textbf{4.82 $\pm$ 0.05} \\
Condition Alignment  & 3.98 $\pm$ 0.44 & 4.51 $\pm$ 0.23 & \textbf{4.55 $\pm$ 0.21} \\ \midrule
\textit{\textbf{Efficiency Metrics}} & & & \\
Params   & 4.3 M             & 43.1 M           & \textbf{3.7 M}           \\
\bottomrule
\end{tabular}
}
\end{table}

To jointly examine extraction quality and system efficiency, Table~\ref{tab:4} reports triplet-quality indicators alongside cost-related metrics. Tri-RAG achieves leading or near-leading scores on logical faithfulness, structural integrity, and condition alignment, while substantially reducing both the average context-token budget and generation-time latency with only a minimal number of trainable parameters. This combined pattern indicates that triplet-structured evidence improves not only the reliability of provided support, but also end-to-end efficiency. By increasing evidence density, compressing redundant context, and limiting the portion of the context window occupied by low-utility retrieved fragments, Tri-RAG yields a more favorable quality–cost trade-off. Conceptually, it uses explicit structure to enable stronger compression and uses better alignment to promote more stable reasoning, which is particularly beneficial under tight long-context budgets or latency-sensitive deployments.

Since Tri-RAG converts unstructured passages into triplets $(T_c, T_p, T_r)$, we evaluate the \emph{intrinsic precision} of structured evidence independently of downstream QA accuracy. Intrinsic precision is operationalized along three complementary dimensions that capture the core requirements of the triplet representation: \emph{faithfulness}, which assesses whether the extracted proof and conclusion are supported by the source passage; \emph{integrity}, which evaluates whether the output conforms to a stable and usable triplet structure with well-defined functional roles; and \emph{alignment}, which measures whether the condition field provides a reliable and compact semantic anchor for query matching. The corresponding results are reported in Table~\ref{tab:4}.

All three metrics are computed using a unified \emph{LLM-as-a-Judge} protocol. The evaluator employs a fixed judging prompt and remains strictly frozen throughout evaluation. To ensure comparability across methods, we use the same evaluation prompt template, input serialization format, and decoding configuration for all settings. The evaluator is constrained to output a discrete integer score on a 5-point scale (1--5), without generating free-form explanations. Deterministic decoding is adopted to eliminate sampling-induced variance. Each instance is evaluated once, and the ``$\pm$'' in Table~\ref{tab:4} denotes the standard deviation across evaluated instances. For each evaluation instance, the judge receives the source passage, the extracted triplet serialized with explicit \texttt{[C]} (Condition), \texttt{[P]} (Proof), and \texttt{[R]} (Conclusion) tags, and, for condition alignment only, the corresponding query $q$. To avoid systematic bias caused by length differences, all inputs are processed using the same formatting and truncation rules prior to evaluation.

\begin{figure}[!t]
    \centering
    \includegraphics[width=1\linewidth]{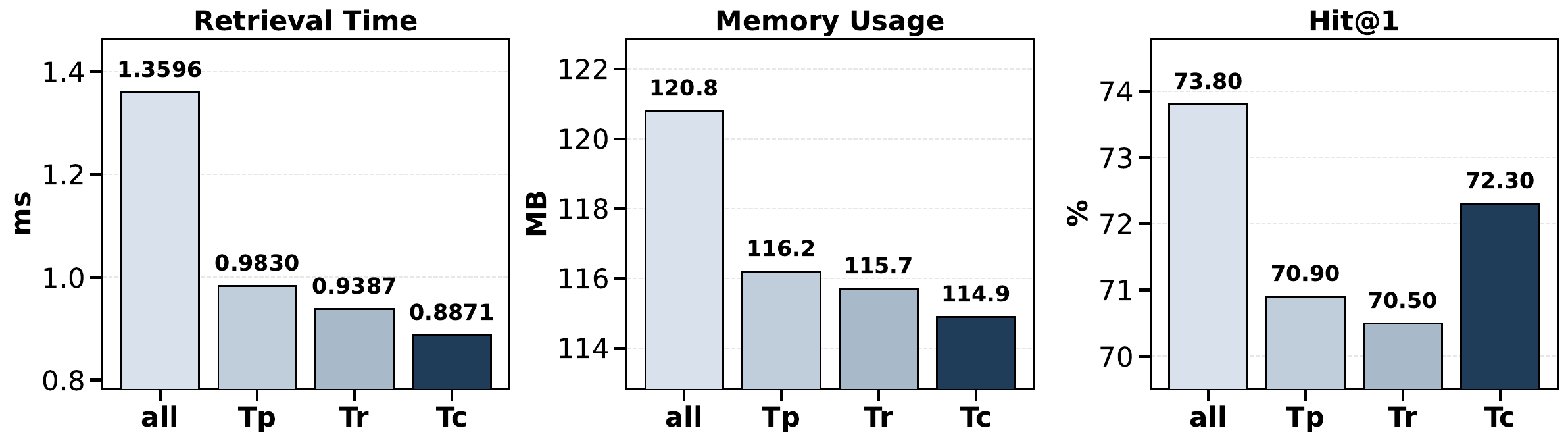}
    \caption{Retrieval-key ablation comparing \textsc{All} retrieval using the concatenation $[T_c;T_p;T_r]$ and single-field retrieval using \textsc{Tc}, \textsc{Tp}, or \textsc{Tr}. We report Hit@1 together with index-level retrieval latency and memory footprint. }
    \label{persistent}
\end{figure}

\subsection{Ablation Study}

To isolate how the \emph{retrieval key} affects matching quality and system cost, we perform a retrieval-key ablation under a fully controlled configuration. Across all settings, we keep the retrieval encoder, index implementation, retrieval depth, and indexing hyperparameters identical, and vary \emph{only} the text used to form the retrieval key. Specifically, we compare \textsc{All}, which retrieves using the concatenation $[T_c;T_p;T_r]$, against three single-field variants that retrieve using only $T_c$ (\textsc{Tc}), $T_p$ (\textsc{Tp}), or $T_r$ (\textsc{Tr}). We report top-1 retrieval hit rate (Hit@1) together with index-level retrieval latency and memory footprint, capturing the trade-off between alignment quality and computational overhead. Hit@1 is a retrieval-layer metric that directly reflects query-key alignment.

The results in Fig.~\ref{persistent} reveal a clear quality--efficiency separation among retrieval keys. All attains the highest Hit@1 of 73.8\%, but also incurs the largest retrieval latency (around 1.360\,ms) and the highest memory footprint (120.8\,MB). 
In contrast, \textsc{Tc} offers the best overall balance, achieving a Hit@1 of 72.3\% while reducing retrieval latency to 0.887\,ms and memory footprint to 114.9\,MB. 
Using \textsc{Tp} or \textsc{Tr} as the retrieval key further decreases Hit@1 to 70.9\% and 70.5\%, respectively, while yielding higher latency and larger memory footprint than \textsc{Tc}. 
Overall, these results suggest that incorporating longer and stylistically diverse fields into the retrieval key does not necessarily improve matching, but can increase both computational and memory overhead.

This behavior is consistent with the functional roles of the triplet fields. $T_c$ is explicitly constructed to encode the applicability scope and constraints of a claim in a compact, query-aligned form, making it a stable and discriminative anchor for retrieval. By contrast, $T_p$ typically contains explanatory language with higher variability in length and expression, which weakens semantic targeting when used as a key. $T_r$ is often concise but lacks the constraint information needed for precise query alignment, making it prone to semantic drift when retrieved in isolation. Taken together, the ablation supports our design choice to anchor retrieval on $T_c$ for efficient and reliable matching, while reserving full-triplet conditioning for downstream reasoning and traceability.

\subsection{Case Studies}
\label{sec:case_studies}

To illustrate the retrieval--reasoning--generation loop of Tri-RAG, we present a representative two-hop example. The query exhibits an explicit dependency chain: it first resolves the relation \emph{work$\rightarrow$author} to obtain an intermediate entity, and then queries \emph{author$\rightarrow$birth year} to produce the final answer. This setup is well-suited for analyzing how early retrieval decisions affect downstream reasoning, since an incorrect intermediate entity typically leads to an irrecoverable failure at the next hop.

\begin{tcolorbox}[
    float,
    floatplacement=h,
    enhanced,
    breakable,
    colback=blue!1,
    colframe=blue!50!black,
    colbacktitle=blue!12!white,
    coltitle=black,
    title={\textbf{Tri-RAG Inference Example}},
    fonttitle=\bfseries\small,
    boxrule=0.75pt,
    arc=2pt,
    left=8pt,
    right=8pt,
    top=6pt,
    bottom=6pt,
    borderline west={3pt}{0pt}{blue!65},
    drop shadow southeast,
    sharp corners
]
\small

\textbf{Source:} 2Wiki\\
\textbf{Question:} What is the birth year of the author of \emph{Harry Potter}?\\
\textbf{Tri-RAG Protocol:} Retrieve by triplet head \texttt{[C]} (Condition), then condition generation on the retrieved full triplet(s) \texttt{[C,P,R]}.\\

\textbf{Tri-RAG Trace:}
\begin{tcolorbox}[
    colback=white,
    colframe=black!25,
    boxrule=0.4pt,
    arc=1.5pt,
    left=6pt,
    right=6pt,
    top=6pt,
    bottom=6pt
]
\ttfamily
\textbf{(Head Retrieval; Hop 1)}\\
Query intent: (Harry\ Potter, author, ?)\quad\textcolor{gray}{\# match against \texttt{[C]} only}\\
Hit A:\ \texttt{[C]} work=\{Harry\ Potter\}; rel=author\ \texttt{[R]} author(Harry\ Potter)=J.K.\ Rowling\\[4pt]

\textbf{(Head Retrieval; Hop 2)}\\
Intermediate: J.K.\ Rowling\\
Query intent: (J.K.\ Rowling, birth\_year, ?)\quad\textcolor{gray}{\# match against \texttt{[C]} only}\\
Hit B:\ \texttt{[C]} subj=\{J.K.\ Rowling\}; attr=birth\_year\ \texttt{[R]} birth\_year(J.K.\ Rowling)=1965\\[6pt]

\textbf{(Full-Triplet Conditioning)}\quad\textcolor{gray}{\# generator sees full structured evidence}\\
Input: $[Q;\ \mathrm{Format}(A,B)]$\\[4pt]

\textbf{Final Answer:} \textcolor{green!40!black}{1965}
\end{tcolorbox}

\end{tcolorbox}

Conventional passage-based RAG retrieves long spans by semantic similarity, which often introduces co-occurrence noise (e.g., multiple entities, several dates, or loosely related biographical facts). Such clutter increases ambiguity during generation and can lead to spurious attribution, especially in multi-hop settings where early errors propagate and contaminate subsequent hops. In addition, passage retrieval typically mixes the matching signal and the generation evidence: the model must both infer which parts of a passage justify an answer and decide whether the passage is even relevant, making reasoning less controllable and harder to audit.

Tri-RAG mitigates this issue via Condition-anchored retrieval. At each hop, matching is restricted to the Condition field, which provides a more regular and tightly scoped retrieval key. In this example, the first hop retrieves triplets aligned with the \emph{work$\rightarrow$author} intent to yield the intermediate entity, and the second hop uses that entity as the next condition anchor to retrieve the birth year evidence. This design reduces retrieval drift and limits low-utility context injection, while enabling a clearer separation between (i) retrieval-time alignment and (ii) generation-time justification.

During generation, Tri-RAG conditions on the full retrieved triplets rather than the matched field alone, presenting premises, supporting rationale, and conclusions in an explicit structured format. This formatting encourages the generator to rely on compact, verifiable evidence rather than incidental cues from verbose passages, improving both reliability and traceability. Importantly, even when multiple candidate entities or dates co-occur in the raw corpus, the structured evidence clarifies which entity is introduced as the intermediate result and which attribute is queried at the next hop, thereby stabilizing multi-step reasoning.

\section{Conclusion}
This paper proposes a structured output method that leverages Soft Prompt Tuning to transform natural language expressions into standardized triplets consisting of a Condition, a Proof, and a Conclusion. By integrating trainable soft prompts into the input, the method enables large language models to generate structured representations without fine-tuning the model’s core parameters. This triplet-based approach provides a robust foundation for downstream tasks such as knowledge retrieval, automated reasoning, and complex question answering. It not only improves the efficiency and accuracy of structure modeling, but also offers a scalable solution for constructing large-scale structured knowledge corpora from unstructured inputs. Empirical results show that our method consistently outperforms rule-based and template-based baselines in both structural consistency and semantic alignment, demonstrating its effectiveness in enhancing reasoning-oriented knowledge extraction. As a next step, we plan to extend the triplet format to accommodate more complex reasoning structures and broader application domains.

\section{Limitations}

Tri-RAG relies on stable structured extraction to ensure reliable retrieval and evidence auditability. In practice, we observe that extracting well-formed triplets remains challenging for certain types of inputs, particularly narrative-heavy passages and cases where key premises or conclusions are distributed across multiple sentences. In such settings, the boundaries between condition, rationale, and conclusion may be less explicit, which can lead to incomplete or ambiguous triplet representations.

Future work could refine Tri-RAG at both extraction and retrieval. On the extraction side, self-verification or consistency checks may help validate triplets against the source text, improving output stability. On the retrieval side, complementary signals (e.g., multi-granular or adaptive representations) could better capture implicit cross-sentence evidence while preserving the triplet schema, improving robustness without sacrificing efficiency or interpretability.

\section*{Acknowledgments}
This work was supported by the National Natural Science Foundation of China (NSFC) under the General Program (Grant No. 62572104).

This article used large language models (such as ChatGPT) as an auxiliary tool in the language polishing process, but did not use them in research conception and academic content generation.

\bibliographystyle{IEEEtran}
\bibliography{reference}
\end{document}